\documentclass[conference]{IEEEtran}
\IEEEoverridecommandlockouts
\usepackage{cite}
\usepackage[T1]{fontenc}
\usepackage{hyperref}
\hypersetup{
    colorlinks=true,
    citecolor=green,
    linkcolor=red,
    urlcolor=blue
}
\usepackage[table]{xcolor}
\usepackage{arydshln}
\usepackage[caption=false,font=footnotesize]{subfig}
\usepackage{amssymb}% http://ctan.org/pkg/amssymb
\usepackage{pifont}% http://ctan.org/pkg/pifont
\newcommand{\xmark}{\ding{55}}%

\usepackage[caption=false,font=footnotesize]{subfig} 
\usepackage{graphicx}
\usepackage{lipsum}
\usepackage{amsmath,amssymb,amsfonts}
\usepackage{xcolor}
\definecolor{c2fColor}{RGB}{157, 91, 82}
\definecolor{CoordAtt}{RGB}{118,144,60}
\definecolor{c3k2Color}{RGB}{56,122,153}
\definecolor{nms}{RGB}{153,153,153}

\usepackage{booktabs}
\usepackage{multirow}
\usepackage{graphicx}
\usepackage{xcolor}

\newcommand{\inc}[1]{\textcolor{blue!60!black}{\scriptsize{$\uparrow$#1}}}

\usepackage{comicneue}

\usepackage{multirow}
\usepackage{booktabs}

\begin{document}

\title{\textbf{KDTwin: }Task-Aware Knowledge Distillation for Lightweight Multi-Task Driving Scene Segmentation}

\author{
\IEEEauthorblockN{
Huy Che\textsuperscript{1,2},
Minh-Khoi Do\textsuperscript{1,2},
Dinh-Duy Phan\textsuperscript{1,2},
and Duc-Khai Lam\textsuperscript{1,2,*}
}
\IEEEauthorblockA{
\textsuperscript{1}University of Information Technology,
Ho Chi Minh City, Vietnam
}
\IEEEauthorblockA{
\textsuperscript{2}Vietnam National University,
Ho Chi Minh City, Vietnam
}
\IEEEauthorblockA{
Email: huycq@uit.edu.vn, khoidm.19@grad.uit.edu.vn,
duypd@uit.edu.vn, khaild@uit.edu.vn
}
\IEEEauthorblockA{
\textsuperscript{*}Corresponding author: Duc-Khai Lam
}
}

\maketitle              % typeset the header of the contribution
% \begingroup\renewcommand\thefootnote{\IEEEauthorrefmark{2}}
% \footnotetext{Equal Contribution}
% \endgroup

\begin{abstract} 
Efficient perception models are essential for real-time autonomous driving, where accuracy and computational cost must be carefully balanced. However, applying knowledge distillation to multi-task driving scene segmentation is challenging because drivable-area and lane segmentation exhibit different spatial characteristics and class imbalance. We propose KDTwin, a task-aware distillation framework for lightweight multi-task segmentation networks. The proposed method performs distillation at both the shared encoder and task-specific decoders. Encoder-level pairwise distillation transfers spatial relational knowledge to enhance the student's shared representation. For the decoders, we use a weighted loss for drivable-area segmentation and a boundary-aware loss for lane segmentation, enabling task-adaptive knowledge transfer without increasing inference complexity. Experiments on BDD100K show consistent improvements across the evaluated CNN-based and Transformer-based student models without increasing inference-time parameters or FLOPs. The results show that designing distillation objectives according to task-specific characteristics can effectively enhance multi-task segmentation performance for autonomous driving. The source code is available at \url{https://github.com/chequanghuy/KDTwin}.
\end{abstract}


\section{Introduction} \label{intro}

Real-time autonomous driving perception requires dense scene understanding under strict computational constraints. In vision-based driving systems, segmentation-based cues provide essential information for downstream decision-making \cite{golf} and motion planning \cite{plaining}. Among them, drivable area segmentation and lane segmentation are two fundamental tasks, as they respectively indicate the safe driving region and the lane structure of the road scene. To improve efficiency, multi-task segmentation networks have been widely adopted to address these two tasks within a unified architecture. Recent lightweight models~\cite{twin, twinmixing, twinplus,mobilevit} have shown promising efficiency for real-time deployment. However, designing a lightweight model often results in noticeable degradation in segmentation accuracy compared to a larger model. For example, in TwinLiteNet$^+$ ~\cite{twinplus}, the large configuration with 1.94M parameters achieves a lane IoU of 34.2\%, while the nano configuration with only 0.03M parameters obtains a lane IoU of 23.3\%. This indicates that although compact models have low computational costs, their performance may still be insufficient for safety-critical driving perception. Therefore, improving the accuracy of lightweight multi-task segmentation models without increasing inference complexity remains an important research problem.

\begin{figure}
    \centering
    \includegraphics[width=0.9\linewidth]{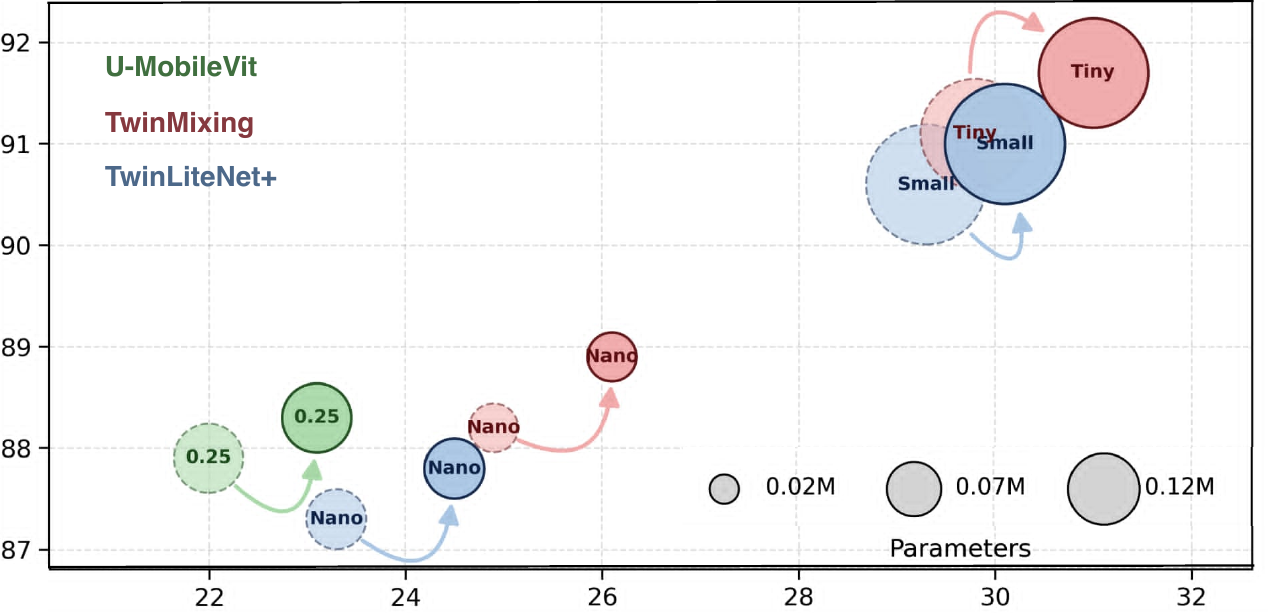}
    \caption{Performance improvement of student models after applying KDTwin. Each point is plotted according to drivable area mIoU on the vertical axis and lane line IoU on the horizontal axis.}
    \label{fig:highlight}
\end{figure}

Model compression techniques \cite{quant,KL,prune} provide an effective way to enhance compact neural networks. Among them, knowledge distillation (KD) is a promising approach that transfers informative knowledge from a high-capacity teacher model to a lightweight student model. Knowledge distillation has been extensively studied in general semantic segmentation \cite{bpkd,structure} and single-task segmentation \cite{st2kd}, where teacher logits \cite{KL,bpkd} or feature representations \cite{structure,adapter} guide the student. Nevertheless, applying knowledge distillation to multi-task driving segmentation remains less explored. A key challenge is that different segmentation tasks exhibit distinct spatial characteristics and class distributions. Drivable area segmentation typically involves large, continuous regions. In contrast, lane segmentation focuses on thin, sparse, and elongated structures, which exhibit severe class imbalance. As a result, using a uniform distillation strategy \cite{KL,structure,adapter} for both tasks may be suboptimal. It can even introduce conflicting supervision between task-specific decoders. This underscores the need for a task-aware distillation strategy that effectively balances shared representation learning and task-specific knowledge transfer.

In this paper, we propose KDTwin, a task-aware knowledge distillation framework for lightweight multi-task driving scene segmentation. KDTwin is designed for two fundamental perception tasks: drivable area segmentation and lane segmentation. To enhance the shared representation of the student model, we introduce encoder-level pairwise distillation that transfers spatial-relational knowledge from the teacher across multiple encoder stages. In addition, we design task-specific decoder distillation objectives according to the characteristics of each task. For drivable area segmentation, we employ a weighted KL-based logit distillation loss to transfer region-level prediction knowledge while reducing the dominance of background pixels. For lane segmentation, we adopt a distillation loss based on BPKD \cite{bpkd} to better preserve the thin, sparse, and boundary-sensitive structure of lane markings. By combining encoder-level relational distillation with task-specific decoder supervision, KDTwin improves compact student models without increasing inference cost.

The main contributions of this paper are summarized as follows: (1) We propose KDTwin, a task-aware knowledge distillation framework for lightweight multi-task driving scene segmentation, targeting both drivable area and lane segmentation. (2) We introduce a hybrid distillation strategy that combines encoder-level pairwise affinity distillation with task-specific decoder objectives, including weighted logit distillation for drivable area segmentation and boundary-aware distillation for lane segmentation. (3) Experiments on the BDD100K dataset \cite{BDD100k} demonstrate that KDTwin consistently improves student models across different architectures without increasing inference complexity. Performance results before and after applying KDTwin are shown in Figure \ref{fig:highlight}.

\section{Related Work} \label{relwork}

\subsection{Lightweight Models for multi-task Segmentation}

Real-world autonomous driving applications \cite{golf} often require operation under strict latency and computational constraints, especially for camera-based perception modules. Among various perception tasks, segmentation plays a crucial role because it provides dense spatial information for downstream decision-making, such as identifying drivable areas and lane markings. Recently, several studies~\cite{twin,twinmixing, twinplus,mobilevit} have focused on developing lightweight multi-task segmentation networks that jointly perform multiple perception tasks within a unified architecture. By leveraging shared representations across tasks, multi-task models can improve inference efficiency and reduce computational cost compared with deploying separate single-task models. However, lightweight models \cite{twinmixing, twinplus, mobilevit} with different configurations also exhibit a clear trade-off between accuracy and computational cost. Although compact configurations are efficient, they often degrade accuracy, particularly on challenging tasks such as lane segmentation. Therefore, improving lightweight multi-task segmentation models without increasing inference cost is important. Knowledge distillation offers a practical solution by transferring knowledge from a high-capacity teacher to a compact student while preserving inference complexity.

\subsection{Knowledge Distillation for Segmentation}

Knowledge distillation has been widely studied for model compression across various vision tasks \cite{KL,adapter,st2kd,bpkd,structure}, including semantic segmentation \cite{bpkd,structure}. Existing segmentation distillation methods can be broadly categorized into feature-based \cite{adapter,structure} and logit-based \cite{KL,bpkd} approaches. Feature-based methods transfer intermediate representations from a high-capacity teacher to a compact student. However, they often require additional adapters \cite{adapter} to align feature dimensions when the teacher and student architectures differ. To alleviate this limitation, relation-based feature distillation methods ~\cite{structure} transfer spatial relationships rather than directly matching raw feature maps. However, computing pairwise affinity matrices on high-resolution feature maps can introduce considerable training overhead. In contrast, logit-based distillation \cite{KL,bpkd} directly aligns the output distributions of the teacher and student, commonly using the KL divergence \cite{KL} to transfer soft prediction knowledge. Although effective for segmentation, KL-based distillation can be sensitive to severe class imbalance, especially when small foreground structures are overwhelmed by dominant background regions. 

However, most existing distillation methods are designed for single-task segmentation, leaving multi-task driving scene segmentation less explored. In this setting, tasks exhibit different spatial characteristics and imbalance patterns, ranging from large, continuous, drivable regions to thin, boundary-sensitive lane structures. This motivates a task-aware distillation framework that combines encoder-level relational distillation with task-specific decoder objectives.

\section{Method}
\begin{figure}[!b]
    \centering
    \includegraphics[width=0.9\linewidth]{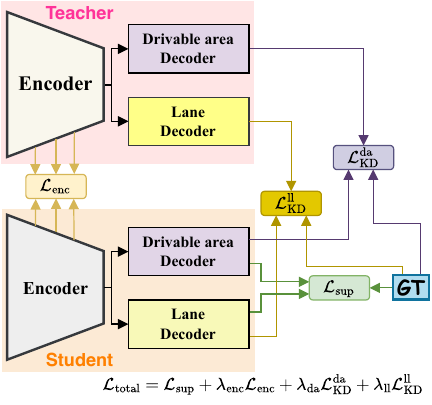}
    \caption{Overview of proposed knowledge distillation methods}
    \label{fig:kdtwin}
\end{figure}

An overview of the proposed KDTwin framework is shown in Figure \ref{fig:kdtwin}, including encoder-level pairwise affinity distillation and task-specific decoder distillation for drivable area and lane segmentation.

\subsection{Encoder-level Pairwise Affinity Distillation}

In multi-task driving scene segmentation, the encoder plays a crucial role in extracting shared representations used by both the drivable area decoder and the lane decoder. Since the student model has a significantly smaller capacity than the teacher model, directly supervising only the final predictions may be insufficient to transfer rich intermediate representations. Therefore, we propose an encoder-level knowledge distillation strategy that aligns the spatial relational structure between student and teacher features across multiple encoder stages via pairwise affinity distillation.

At each encoder stage $l$, we obtain the feature maps of the student and teacher, denoted as $\mathbf{F}^{l}_{s} \in \mathbb{R}^{C^{l}_{s} \times H^{l}_{s} \times W^{l}_{s}}$ and $\mathbf{F}^{l}_{t} \in \mathbb{R}^{C^{l}_{s} \times H^{l}_{s} \times W^{l}_{s}}$, respectively. To enable the student to learn the spatial relational structure from the teacher, we construct an affinity matrix $\mathbf{A}^{l}$ to measure the similarity between spatial positions within each feature map. However, directly computing the affinity matrix on high-resolution feature maps incurs significant memory and computational costs, especially in the early stages of the encoder, when the spatial resolution is high. Therefore, before computing the affinity matrix, we apply max pooling to downsample the feature maps:

\begin{equation}
\bar{\mathbf{F}}^{l}_{s}
=
\operatorname{Pool}
\left(
\mathbf{F}^{l}_{s}
\right),
\quad
\bar{\mathbf{F}}^{l}_{t}
=
\operatorname{Pool}
\left(
\mathbf{F}^{l}_{t}
\right),
\end{equation}

After pooling, the feature maps are reduced to the spatial size $H' \times W'$. Although downsampling may remove some local details, neighbouring responses in early high-resolution feature maps are usually highly correlated. Thus, reducing the spatial resolution can significantly decrease memory consumption and computational cost during training while still preserving global spatial relational information.

The pooled feature maps are then reshaped into spatial feature sequences:

\begin{equation}
\bar{\mathbf{F}}^{l}_{s},
\bar{\mathbf{F}}^{l}_{t}
\in
\mathbb{R}^{C_l \times N'},
\quad
N' = H' \times W'.
\end{equation}

The pairwise affinity matrix is computed based on cosine similarity between spatial feature representations:

\begin{equation}
\mathbf{A}^{l}
=
\frac{
\left(\bar{\mathbf{F}}^{l}\right)^{\top}
\bar{\mathbf{F}}^{l}
}{
\left\|\bar{\mathbf{F}}^{l}\right\|_2
\left\|\bar{\mathbf{F}}^{l}\right\|_2
},
\quad
\mathbf{A}^{l}
\in
\mathbb{R}^{N' \times N'}.
\end{equation}

Here, $\mathbf{A}^{l}$ represents the spatial affinity matrix of the feature map at stage $l$. Each element in $\mathbf{A}^{l}$ indicates the cosine similarity between a pair of spatial positions in the feature map. In practice, this operation is implemented by applying $\ell_2$ normalization along the channel dimension before matrix multiplication. Accordingly, we obtain the student and teacher affinity matrices, denoted as $\mathbf{A}^{l}_{s}$ and $\mathbf{A}^{l}_{t}$, respectively.

We use the mean squared error of the student and teacher affinity matrices to formulate the pairwise affinity distillation loss at stage $l$:

\begin{equation}
\mathcal{L}^{l}_{\text{PA}}
=
\operatorname{MSE}(\mathbf{A}^{l}_s,\mathbf{A}^{l}_t)
\end{equation}

Finally, the overall encoder-level distillation loss is computed as the weighted sum of pairwise affinity losses over the selected encoder stages:

\begin{equation}
\mathcal{L}_{\text{enc}}
=
\sum_{l=1}^{L}
\lambda^{l}_{\text{enc}}
\mathcal{L}^{l}_{\text{PA}},
\end{equation}

where $\lambda^{l}_{\text{enc}}$ controls the contribution of the $l$-th encoder stage. By aligning the spatial-relational structure between teacher and student features, this loss encourages the student encoder to learn more informative shared representations for both segmentation tasks.

\subsection{Task-specific Decoder Knowledge Distillation}
Although the two segmentation tasks share the same encoder, their decoder branches focus on different spatial characteristics. Drivable area segmentation primarily involves large, continuous regions, whereas lane segmentation requires preserving thin, sparse, and boundary-sensitive structures. Therefore, applying the same distillation objective to both decoders may be suboptimal. To address this issue, we design task-specific decoder distillation losses for the drivable area and lane branches.

\subsubsection{Drivable Area Decoder Distillation}

For the drivable area decoder, we adopt a weighted logit distillation loss to transfer region-level prediction knowledge from the teacher to the student. Let $\mathbf{z}^{da}_{s}$ and $\mathbf{z}^{da}_{t}$ denote the student and teacher logits for drivable area segmentation. The temperature $\tau$ is introduced to soften the teacher and student probability distributions, allowing the student to learn richer inter-class information from the teacher.

\begin{equation}
\mathbf{D}^{da}
=
\tau^{2}
\operatorname{KL}
\left(
\operatorname{softmax}(\mathbf{z}^{da}_{t}/{\tau})
\;\middle\|\;
\operatorname{softmax}(\mathbf{z}^{da}_{s}/\tau)
\right),
\end{equation}

\begin{figure}[!b]
    \centering
    \includegraphics[width=\linewidth]{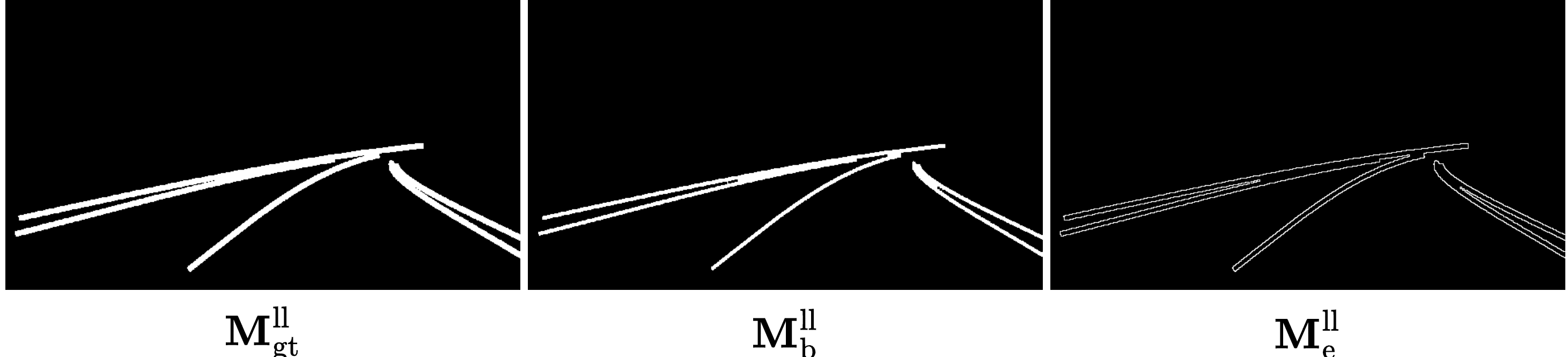}
    \caption{Illustration of the lane ground-truth, body, and edge masks}
    \label{fig:lane}
\end{figure}

Since the background region usually occupies a larger area than the foreground drivable region, the pixel-wise distillation loss can be dominated by background pixels. For example, in many BDD100K images, the drivable foreground occupies a smaller area than the background. This imbalance may reduce the contribution of the drivable area during knowledge transfer. To alleviate this issue, we construct a spatial weight map from the ground-truth drivable area mask $\mathbf{M}^{da}$:
\begin{equation}
\mathbf{W}^{da}
=
1
+
(\gamma_{da}-1)
\mathbf{M}^{da},
\end{equation}

where $\gamma_{da}$ denotes the foreground weighting factor. The final drivable area distillation loss is defined as:

\begin{equation}
\mathcal{L}^\text{da}_{\text{KD}}
=
\frac{\left\|
\mathbf{W}^{da}
\odot
\mathbf{D}^{da}
\right\|_1}{
\left\|
\mathbf{W}^{da}
\right\|_1}
\end{equation}

This objective enables the student decoder to learn the teacher's softened prediction distribution while giving proper emphasis to drivable area regions.

\subsubsection{Lane Decoder Distillation}

Lane markings are thin, sparse, and highly sensitive to boundary localization. Therefore, pixel-wise logit distillation can be dominated by background pixels, providing insufficient supervision for lane boundaries. As shown in the first subfigure in Figure~\ref{fig:lane}, lane pixels occupy only a small portion of the image, approximately 5\% of the image area in BDD100K~\cite{BDD100k}. To mitigate this imbalance, we adopt a boundary-aware partitioned distillation loss inspired by BPKD~\cite{bpkd}. This design separates the distillation process into edge and body regions, encouraging the student to focus on boundary-sensitive lane structures while maintaining prediction consistency in non-edge areas.

Given the ground-truth lane mask $\mathbf{M}^{ll}$, we generate an edge mask $\mathbf{M}^{ll}_{e}$ using a local morphological operation:
$\mathbf{M}^{ll}_{e}=\operatorname{dilation}(\mathbf{M}^{ll})-\operatorname{erosion}(\mathbf{M}^{ll}).$
The body mask is defined as the complementary region:
$
\mathbf{M}^{ll}_{b}
=
1
-
\mathbf{M}^{ll}_{e}$. The edge and body masks are visualized in Figure~\ref{fig:lane}. Let $\mathbf{z}^{ll}_{s}$ and $\mathbf{z}^{ll}_{t}$ denote the student and teacher logits for lane segmentation, respectively. The masked logits for the edge and body regions are obtained as:
\begin{equation}
\tilde{\mathbf{z}}^{e}_{s,t}
=
\mathbf{z}^{ll}_{s,t}
\odot
\mathbf{M}^{ll}_{e},
\quad
\tilde{\mathbf{z}}^{b}_{s,t}
=
\mathbf{z}^{ll}_{s,t}
\odot
\mathbf{M}^{ll}_{b},
\end{equation}
where $\odot$ denotes element-wise multiplication.

We then apply KL divergence to both the edge and body logits, encouraging the student to mimic the teacher's softened prediction distribution:
% \begin{equation}
% \mathbf{D}^{ll}_{e}
% =
% \operatorname{KL}
% \left(
% \tilde{\mathbf{z}}^{e}_{t}
% \;\middle\|\;
% \tilde{\mathbf{z}}^{e}_{s}
% \right),\quad
% \mathbf{D}^{ll}_{b}
% =
% \operatorname{KL}
% \left(
% \tilde{\mathbf{z}}^{b}_{t}
% \;\middle\|\;
% \tilde{\mathbf{z}}^{b}_{s}
% \right)
% \end{equation}

\begin{equation}
\begin{split}
\mathbf{D}^{ll}_{e}
&=
\operatorname{KL}
\left(
\operatorname{softmax}(
\tilde{\mathbf{z}}^{e}_{t}
)
\;\middle\|\;
\operatorname{softmax}(
\tilde{\mathbf{z}}^{e}_{s}
)
\right),\\
\mathbf{D}^{ll}_{b}
&=
\operatorname{KL}
\left(
\operatorname{softmax}(
\tilde{\mathbf{z}}^{b}_{t}
)
\;\middle\|\;
\operatorname{softmax}(
\tilde{\mathbf{z}}^{b}_{s}
)
\right).
\end{split}
\end{equation}

The edge and body distillation losses are normalized by the corresponding valid mask regions:
\begin{equation}
\mathcal{L}^\text{ll}_{e}
=
\frac{
\left\|
\mathbf{M}^{ll}_{e}
\odot
\mathbf{D}^{ll}_{e}
\right\|_{1}
}{
\left\|
\mathbf{M}^{ll}_{e}
\right\|_{1}
},\quad
\mathcal{L}^\text{ll}_{b}
=
\frac{
\left\|
\mathbf{M}^{ll}_{b}
\odot
\mathbf{D}^{ll}_{b}
\right\|_{1}
}{
\left\|
\mathbf{M}^{ll}_{b}
\right\|_{1}
}
\end{equation}

Finally, the lane decoder distillation loss is defined as a weighted combination of the body and edge terms:
\begin{equation}
\mathcal{L}^\text{ll}_{\text{KD}}
=
\lambda_{\text{body}}
\mathcal{L}^\text{ll}_{b}
+
\lambda_{\text{edge}}
\mathcal{L}^\text{ll}_{e},
\end{equation}
where $\lambda_{\text{body}}$ and $\lambda_{\text{edge}}$ control the contributions of the body and edge regions, respectively. The edge term strengthens knowledge transfer around lane boundaries, while the body term preserves prediction consistency in non-edge regions. This formulation is well suited for lane segmentation, where small boundary shifts can substantially reduce IoU due to the thin structure of lane markings.

\subsection{Overall Training Objective}

For each architecture, $\mathcal{L}_{\text{sup}}$ is the original supervised segmentation objective \cite{focal,tversky} used to train its baseline student.

The overall training objective is defined as:
\begin{equation}
\mathcal{L}_{\text{total}}
=
\mathcal{L}_{\text{sup}}
+
\lambda_{\text{enc}}\mathcal{L}_{\text{enc}}
+
\lambda_{da}\mathcal{L}^\text{da}_{\text{KD}}
+
\lambda_{ll}\mathcal{L}^\text{ll}_{\text{KD}},
\end{equation}
where $\mathcal{L}_{\text{enc}}$ denotes the encoder-level pairwise affinity distillation loss, while $\mathcal{L}^\text{da}_{\text{KD}}$ and $\mathcal{L}^\text{ll}_{\text{KD}}$ denote the task-specific decoder distillation losses for drivable area segmentation and lane segmentation, respectively. The coefficients $\lambda_{\text{enc}}$, $\lambda_{da}$, and $\lambda_{ll}$ control the contribution of each distillation term.

\begin{figure}[!t]
    \centering
    \includegraphics[width=0.9\linewidth]{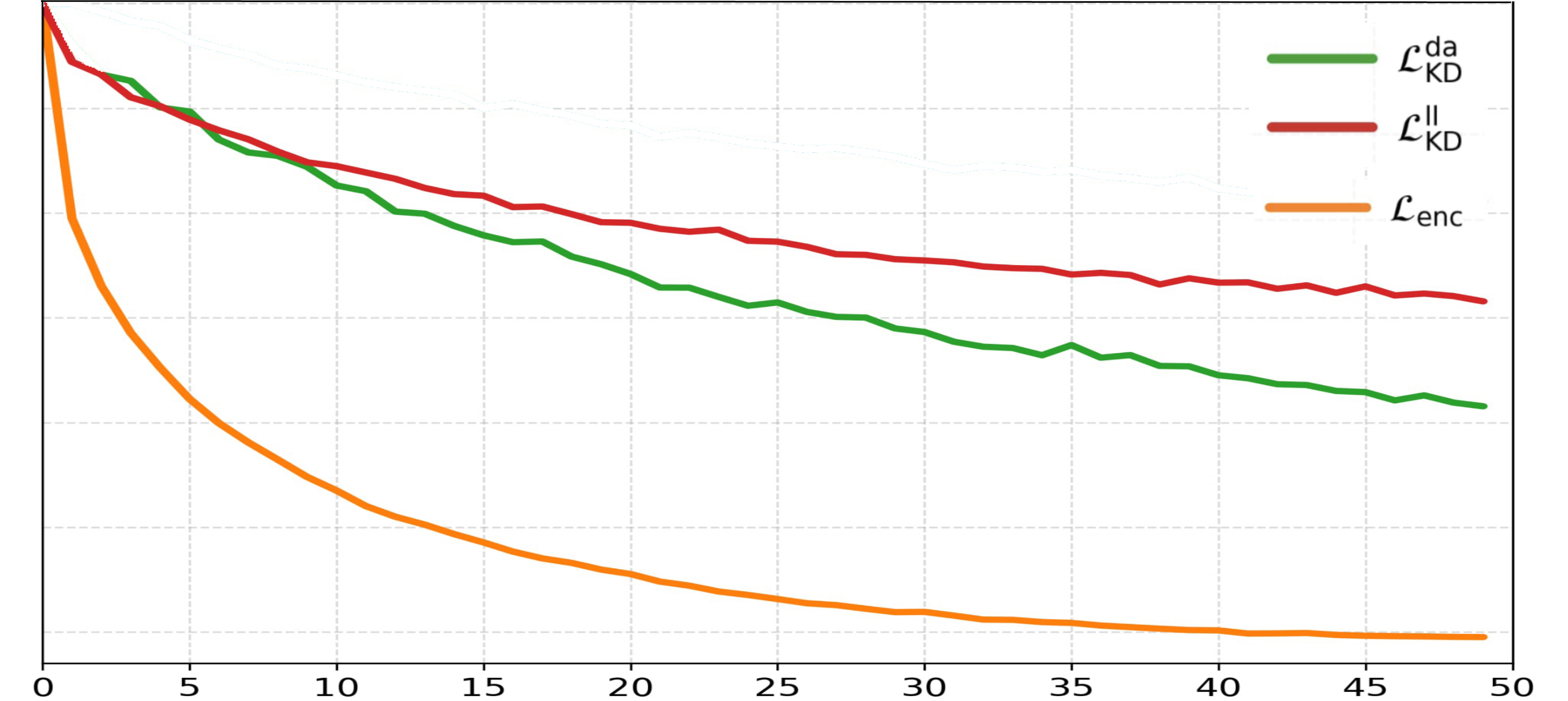}
    \caption{Evolution of normalized knowledge distillation losses over 50 epochs.}
    \label{fig:loss}
\end{figure}
\section{Results}

\subsection{Experimental Settings}

We conduct experiments on the BDD100K dataset \cite{BDD100k}, which comprises 70K training images and 10K validation images. Rather than training student models from scratch, we initialize them with pretrained weights and train them with the proposed knowledge distillation framework for 50 epochs. To ensure a fair comparison, we keep each student model's supervised training configuration unchanged from its original baseline. For evaluation, we report mIoU for drivable area segmentation, and accuracy and IoU for lane segmentation. We also report the number of parameters and FLOPs to measure model complexity and computational cost.

We set the weights for drivable area and lane logit distillation to $\lambda_{da}=0.65$ and $\lambda_{ll}=0.35$, respectively. For the weighted KL distillation in the drivable area decoder, the temperature and foreground weighting factor are set to $\tau_{da}=2.0$ and $\gamma_{da}=1.5$. For the boundary-aware lane decoder distillation, we set $\lambda_{\text{body}}=0.2$, and $\lambda_{\text{edge}}=0.8$. For encoder-level pairwise affinity distillation, the weights of the three selected encoder stages are set to $\lambda^{1, 2, 3}_{\text{enc}}=\{0.05, 0.05, 0.5\}$.

\subsection{Main results}

\begin{figure}[!b]
    \centering
    \includegraphics[width=\linewidth]{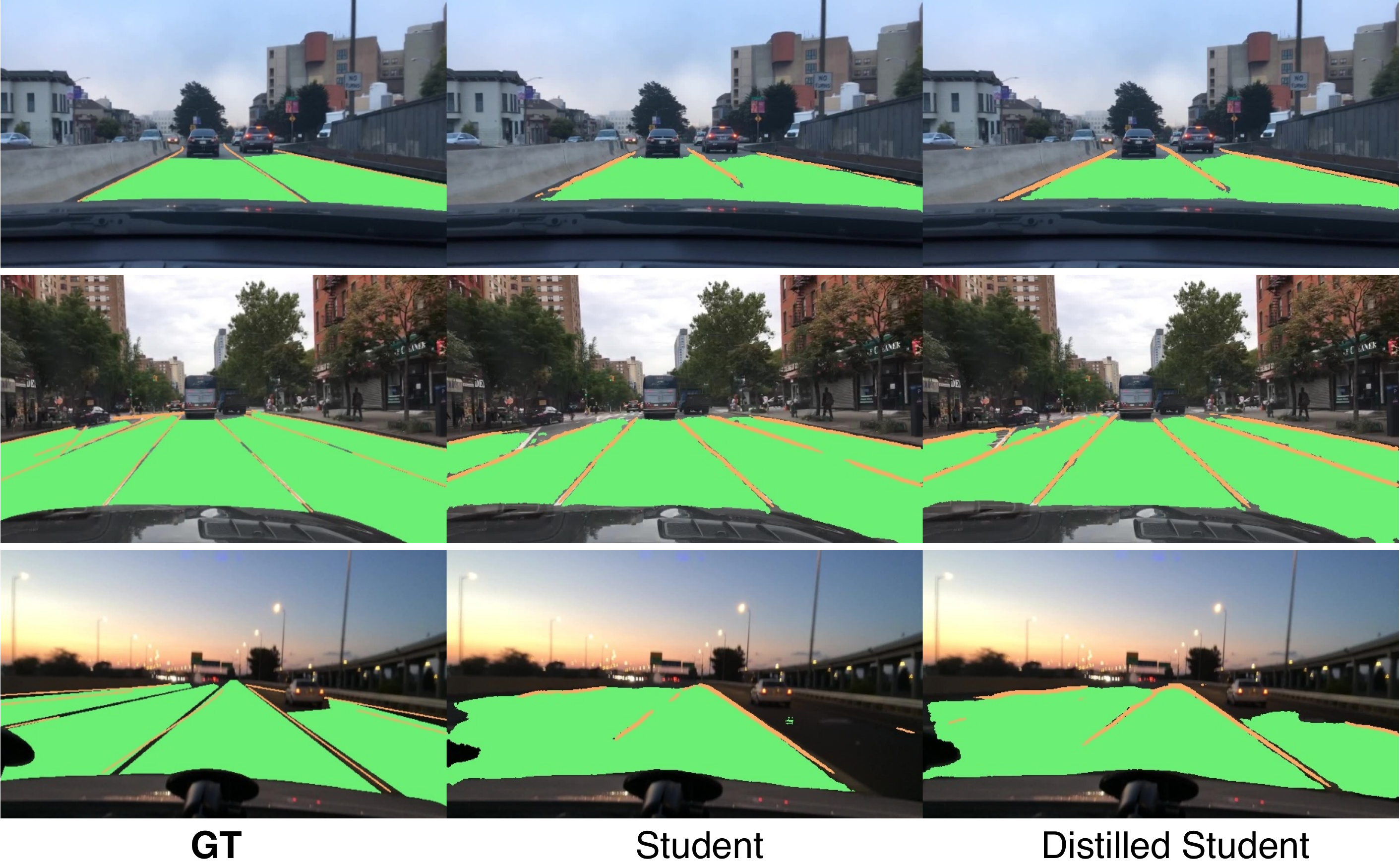}
    \caption{Qualitative results for TwinMixing$_\text{Nano}$}
    \label{fig:vis}
\end{figure}

\begin{table*}[h]
\centering
\caption{Performance comparison of teacher, student, and distilled student models on BDD100K.}
\label{tab:results}
\setlength{\tabcolsep}{12pt}
\begin{tabular}{lccccccc}
\toprule
\multirow{2}{*}{\textbf{Model}} & \textbf{Drivable area segmentation} &  & \multicolumn{2}{c}{\textbf{Lane segmentation}} & \multirow{2}{*}{\textbf{FLOPs}} & \multirow{2}{*}{\textbf{Parameters}} & \multicolumn{1}{l}{\multirow{2}{*}{$\bar{\Delta}$}} \\ 
\cmidrule{2-2} \cmidrule{4-5}
 & \textbf{mIoU (\%)} &  & \textbf{Acc (\%)} & \textbf{IoU (\%)} &  &  & \multicolumn{1}{l}{} \\ 
\midrule

\multicolumn{8}{l}{\textit{\textbf{TwinLiteNet$^+$}}} \\ 
\midrule
\rowcolor{gray!25}Teacher: TwinLiteNet$^+_{\text{Large}}$ & 92.9 &  & 81.9 & 34.2 & 17.58G & 1.94M &  \\ 
\midrule
Student: TwinLiteNet$^+_{\text{Nano}}$ & 87.3 &  & 70.2 & 23.3 & \multirow{2}{*}{0.57G} & \multirow{2}{*}{0.03M} & \multirow{2}{*}{+2.1} \\
Distilled Student & 87.8 \inc{0.5} &  & 74.8 \inc{4.6} & 24.5 \inc{1.2} &  &  &  \\ 
\midrule
Student: TwinLiteNet$^+_{\text{Small}}$ & 90.6 &  & 75.8 & 29.3 & \multirow{2}{*}{1.40G} & \multirow{2}{*}{0.12M} & \multirow{2}{*}{+1.5} \\
Distilled Student & 91.0 \inc{0.4} &  & 79.0 \inc{3.2} & 30.1 \inc{0.8} &  &  &  \\ 
\midrule

\multicolumn{8}{l}{\textit{\textbf{TwinMixing}}} \\ 
\midrule
\rowcolor{gray!25}Teacher: TwinMixing$_{\text{Large}}$ & 92.8 &  & 82.4 & 34.2 & 14.25G & 1.50M &  \\ 
\midrule
Student: TwinMixing$_{\text{Nano}}$ & 88.2 &  & 71.8 & 24.9 & \multirow{2}{*}{0.44G} & \multirow{2}{*}{0.02M} & \multirow{2}{*}{+1.9} \\
Distilled Student & 88.9 \inc{0.7} &  & 75.7 \inc{3.9} & 26.1 \inc{1.2} &  &  &  \\ 
\midrule
Student: TwinMixing$_{\text{Tiny}}$ & 91.1 &  & 76.6 & 29.8 & \multirow{2}{*}{1.08G} & \multirow{2}{*}{0.10M} & \multirow{2}{*}{+1.5} \\
Distilled Student & 91.7 \inc{0.6} &  & 79.2 \inc{2.6} & 31.0 \inc{1.2} &  &  &  \\ 
\midrule

\multicolumn{8}{l}{\textit{\textbf{U-MobileViT}}} \\ 
\midrule
\rowcolor{gray!25}Teacher: U-MobileViT 2.0 & 92.0 &  & 78.8 & 32.0 & 3.47G & 2.25M &  \\ 
\midrule
Student: U-MobileViT 0.25 & 87.9 &  & 68.5 & 22.0 & \multirow{2}{*}{0.12G} & \multirow{2}{*}{0.04M} & \multirow{2}{*}{+1.4} \\
Distilled Student & 88.3 \inc{0.4} &  & 71.2 \inc{2.7} & 23.1 \inc{1.1} &  &  &  \\ 
\bottomrule
\end{tabular}%
\vspace{2pt}
\parbox{\textwidth}{
\footnotesize
\textit{Note:} The results of Student: TwinMixing$_{\text{Nano}}$, Teacher: U-MobileViT 2.0, and Student: U-MobileViT 0.25 are reproduced using the publicly provided source code. $\bar{\Delta}$ denotes the average improvement over the three main metrics: mIoU, acc, and IoU.
}
\end{table*}

To demonstrate the effectiveness of the proposed knowledge distillation method for multi-task segmentation in autonomous driving scenarios, we evaluate it on different model families, including CNN-based architectures such as TwinLiteNet$^+$ \cite{twinplus} and TwinMixing \cite{twinmixing}, as well as a Vision Transformer-based architecture, U-MobileViT \cite{mobilevit}. The results in Table~\ref{tab:results} show that the proposed method consistently improves the performance of student models on both drivable area segmentation and lane segmentation while preserving the same number of parameters and FLOPs during inference.

Across different student models, the proposed distillation framework consistently improves segmentation performance without increasing inference cost. The positive $\bar{\Delta}$ values, ranging from $+1.4$ to $+2.1$, indicate that the method generalizes across both CNN-based and Transformer-based lightweight architectures. Notably, the gains are more pronounced for lane segmentation, with lane accuracy improved by up to $4.6$ and lane IoU by up to $1.2$. Thin and sparse lane markings make pixel-level predictions highly sensitive to small boundary errors, which explains the larger improvement in lane accuracy. Moreover, the student-teacher gap is larger for lane segmentation than for drivable area segmentation, especially in IoU, indicating that compact students still struggle to capture reliable lane representations.
Nevertheless, the proposed method consistently improves both tasks, indicating that task-specific distillation can enhance lane segmentation while still benefiting drivable area segmentation. Figure 4 shows that the proposed distillation losses decrease consistently during training, indicating stable optimization of the student model. Furthermore, the qualitative results shown in Figure \ref{fig:vis} also demonstrate the effectiveness of the proposed method.

\subsection{Ablation study}

\subsubsection{Ablation study on distillation components}

Table~\ref{tab:ablation_components} analyzes the contribution of each distillation component. Removing the encoder-level distillation loss $\mathcal{L}_{\text{enc}}$ degrades performance on both tasks, indicating that pairwise affinity distillation helps improve the shared representation used by the two decoders. Removing $\mathcal{L}^\text{da}_{\text{KD}}$ mainly affects drivable area segmentation, while removing $\mathcal{L}^\text{ll}_{\text{KD}}$ leads to a clear drop in lane accuracy and IoU. These results show that each task-specific decoder distillation loss contributes to its corresponding task, and the full objective provides the most balanced performance across drivable area and lane segmentation.

\begin{table}[]
\caption{Ablation study on the proposed distillation components.}
\label{tab:ablation_components}
\begin{tabular}{lcccc}
\toprule
\multirow{2}{*}{\textbf{}}              & \textbf{Drivable area} &                      & \multicolumn{2}{c}{\textbf{Lane}}           \\ \cline{2-2} \cline{4-5} 
                                        & \textbf{mIoU (\%)}     &                      & \textbf{Acc (\%)}    & \textbf{IoU (\%)}    \\ \midrule
Teacher: TwinMixing$_{\text{Large}}$ & 92.8                    &                      & 82.4 & 34.2                 \\
Student: TwinMixing$_{\text{Nano}}$   & 88.2 &  & 71.8 & 24.9                 \\
Distilled Student                       & 88.9                   &                      & 75.7                 & 26.1                 \\ \midrule
\rowcolor{gray!25}w/o $\mathcal{L}_{\text{enc}}$          &         88.5               &                      &           75.0           &          25.5            \\
\addlinespace[2pt]
\rowcolor{gray!25}w/o $\mathcal{L}^\text{da}_{\text{KD}}$      &        88.1                &                      &        75.3              &         25.8             \\
\addlinespace[2pt]
\rowcolor{gray!25}w/o $\mathcal{L}^\text{ll}_{\text{KD}}$      & 89.0   & \multicolumn{1}{l}{} & 71.4 & 24.4 \\ \bottomrule
\end{tabular}%
% }
\end{table}

\subsubsection{Effect of Task-specific Decoder Training}

Table~\ref{tab:only_decoder} shows the effect of training each task-specific decoder independently. The ``Only'' setting means that the model is optimized only for the corresponding decoder. Compared with the jointly distilled student, single-decoder training achieves slightly higher performance on its target task, increasing drivable area mIoU from $88.9\%$ to $89.7\%$ and lane IoU from $26.1\%$ to $26.5\%$. This indicates that optimizing a single decoder avoids the trade-off introduced by jointly learning multiple tasks. However, the jointly distilled student still improves both tasks over the original student, showing that the proposed framework achieves a practical balance between task-specific optimization and multi-task performance.

\begin{table}[]
\caption{Effect of single-decoder training on task-specific segmentation performance.}
\label{tab:only_decoder}
\begin{tabular}{lcccc}
\toprule
\multirow{2}{*}{\textbf{}}              & \textbf{Drivable area} &  & \multicolumn{2}{c}{\textbf{Lane}}     \\ \cmidrule{2-2} \cmidrule{4-5} 
                                        & \textbf{mIoU (\%)}     &  & \textbf{Acc (\%)} & \textbf{IoU (\%)} \\ \midrule
Teacher: TwinMixing$_{\text{Large}}$ & 92.8                    &                      & 82.4 & 34.2                 \\
Student: TwinMixing$_{\text{Nano}}$   & 88.2 &  & 71.8 & 24.9                 \\
\rowcolor{gray!15}Distilled Student                       & 88.9                   &  & 75.7              & 26.1              \\ \midrule
\rowcolor{gray!25}\textit{Only} drivable area decoder            & 89.7                   &  & \xmark                 & \xmark                 \\
\rowcolor{gray!25}\textit{Only} lane decoder                     & \xmark                      &  & 76.3              & 26.5              \\ \bottomrule
\end{tabular}%
\end{table}

\subsubsection{Effect of different teacher architectures}

Table~\ref{tab:different_teachers} evaluates the effect of using different teacher architectures for distilling TwinMixing$_{\text{Nano}}$. Both TwinMixing$_{\text{Large}}$ and TwinLiteNet$^+_{\text{Large}}$ improve the student over the supervised baseline, showing that the proposed method can benefit from different high-capacity teachers. The teacher from the same model family, TwinMixing$_{\text{Large}}$, achieves slightly better results, suggesting that architectural consistency can facilitate knowledge transfer. Nevertheless, the gains obtained with TwinLiteNet$^+_{\text{Large}}$ indicate that the proposed framework can also work with heterogeneous teacher-student pairs.

\section{Discussion and Conclusion}
\subsection{Limitation}

The proposed framework has two main limitations. First, the hyperparameters and loss weights are empirically selected and kept fixed during training, which may not be optimal across different architectures, student capacities, or task difficulties. Second, encoder-level pairwise affinity distillation introduces additional training overhead due to the computation of spatial affinity matrices. However, spatial pooling helps reduce this cost.
\subsection{Conclusion}
In this paper, we propose KDTwin, a task-aware knowledge distillation framework for lightweight multi-task driving scene segmentation. The proposed method combines encoder-level pairwise affinity distillation with task-specific decoder distillation for drivable area and lane segmentation. Experiments on BDD100K demonstrate consistent improvements across CNN-based and Transformer-based student models without increasing inference cost. Future work will explore additional adaptive weighting strategies to balance task-specific distillation objectives.

\begin{table}[]
\centering
\caption{Effect of different teacher architectures on the distilled student.}
\label{tab:different_teachers}
\begin{tabular}{lcccc}
\toprule
\multirow{2}{*}{\textbf{Model}} 
& \textbf{Drivable area} &  & \multicolumn{2}{c}{\textbf{Lane}} \\ 
\cmidrule{2-2} \cmidrule{4-5} 
& \textbf{mIoU (\%)} &  & \textbf{Acc (\%)} & \textbf{IoU (\%)} \\ 
\midrule
\rowcolor{gray!15}Student: TwinMixing$_{\text{Nano}}$     & 88.2 &  & 71.8 & 24.9 \\ 
\midrule
Teacher: TwinMixing$_{\text{Large}}$    & 92.8 &  & 82.4 & 34.2 \\
\rowcolor{gray!25}
Distilled Student                       & 88.9 &  & 75.7 & 26.1 \\ 
\midrule
Teacher: TwinLiteNet$^+_{\text{Large}}$ & 92.9 &  & 81.9 & 34.2 \\
\rowcolor{gray!25}
Distilled Student                       & 88.8 &  & 74.3 & 25.9 \\ 
\bottomrule
\end{tabular}
\end{table}

\section{ACKNOWLEDGMENT}
This research is funded by University of Information Technology-Vietnam National University of Ho Chi Minh city under grant number CS4-2026-80038.

{\small
\bibliographystyle{IEEEtran}
\bibliography{ref}
}

\end{document}